\documentclass[sigconf]{acmart}
\AtBeginDocument{%
  }

\setcopyright{acmlicensed}
\copyrightyear{2018}
\acmYear{2018}
\acmDOI{XXXXXXX.XXXXXXX}
\acmConference[Conference acronym 'XX]{Make sure to enter the correct
  conference title from your rights confirmation email}{June 03--05,
  2018}{Woodstock, NY}
\acmISBN{978-1-4503-XXXX-X/2018/06}

\usepackage{multirow}
\usepackage{enumitem}

\copyrightyear{2026}
\acmYear{2026}
\setcopyright{cc}
\setcctype{by}
\acmConference[MM '26]{Proceedings of the 34th ACM International Conference on Multimedia}{November 10--14, 2026}{Rio de Janeiro, Brazil}
\acmBooktitle{Proceedings of the 34th ACM International Conference on Multimedia (MM '26), November 10--14, 2026, Rio de Janeiro, Brazil}
\acmDOI{10.1145/3767308.3834951}
\acmISBN{979-8-4007-2213-4/2026/11}

\begin{document}

\title{ArtifactWorld: Scaling 3D Gaussian Splatting Artifact Restoration via Video Generation Models}

\author{Xinliang Wang}
\authornote{Equal contribution.\quad
  \textsuperscript{\textdagger}Corresponding author.}
\affiliation{%
  \institution{KE Holdings Inc.}
  \city{Beijing}
  \country{China}}
\email{wangxinliang@buaa.edu.cn}

\author{Yifeng Shi}
\authornotemark[1] 
\authornotemark[2] 
\affiliation{%
  \institution{KE Holdings Inc.}
  \city{Beijing}
  \country{China}}
\email{shiyifeng@tju.edu.cn}

\author{Zhenyu Wu}
\affiliation{%
  \institution{KE Holdings Inc.}
  \city{Beijing}
  \country{China}}
\email{zhenyuwu414@gmail.com}

\renewcommand{\shortauthors}{Xinliang Wang, Yifeng Shi, and Zhenyu Wu}

\begin{abstract}
  3D Gaussian Splatting (3DGS) delivers high-fidelity real-time rendering but suffers from geometric and photometric degradations under sparse-view constraints. Current generative restoration approaches are often limited by insufficient temporal coherence, a lack of explicit spatial constraints, and a lack of large-scale training data, resulting in multi-view inconsistencies, erroneous geometric hallucinations, and limited generalization to diverse real-world artifact distributions. In this paper, we present ArtifactWorld, a framework that resolves 3DGS artifact repair through systematic data expansion and a homogeneous dual-model paradigm. To address the data bottleneck, we establish a fine-grained phenomenological taxonomy of 3DGS artifacts and construct a comprehensive training set of 107.5K diverse paired video clips to enhance model robustness. Architecturally, we unify the restoration process within a video diffusion backbone, utilizing an isomorphic predictor to localize structural defects via an artifact heatmap. This heatmap then guides the restoration through an Artifact-Aware Triplet Fusion mechanism, enabling precise, intensity-guided spatio-temporal repair within native self-attention. Extensive experiments demonstrate that ArtifactWorld achieves state-of-the-art performance in sparse novel view synthesis and robust 3D reconstruction. Code available at: \href{https://github.com/fyting/ArtifactWorld}{this https URL}.
\end{abstract}


\begin{teaserfigure}
  \centering
  \includegraphics[width=\textwidth]{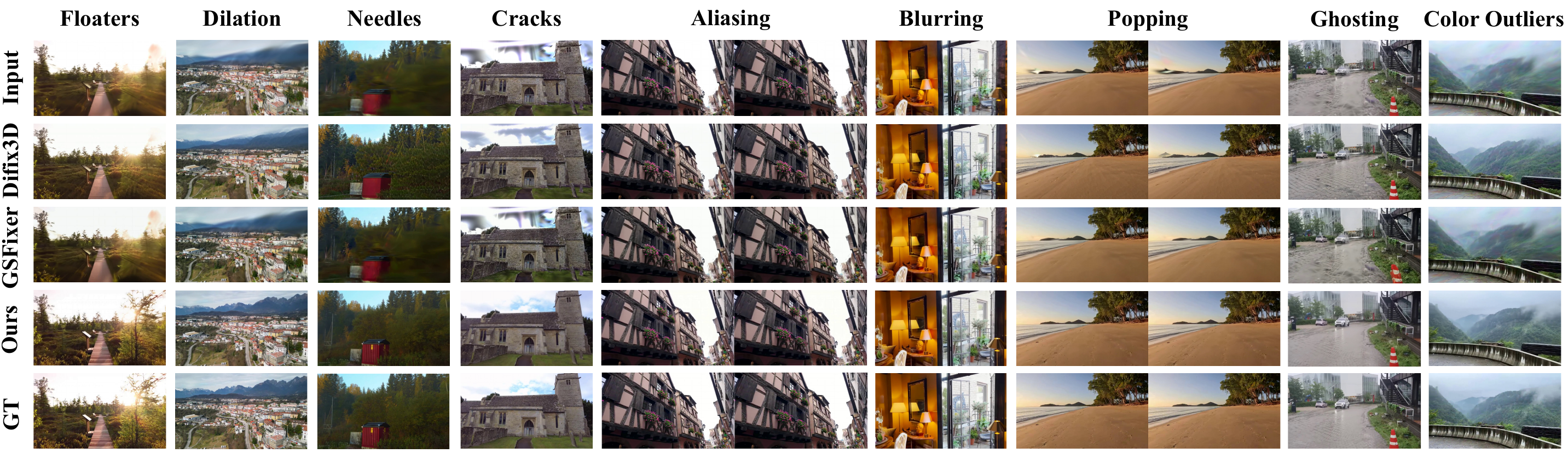}
  \vspace{-5mm}
  \caption{ArtifactWorld effectively resolves complex 3D Gaussian Splatting degradations under sparse-view constraints. We systematically categorize these failures into nine fine-grained phenomenological artifact types (Sec.~\ref{sec:data}). Compared to recent generative restoration pipelines (e.g., Difix3D~\cite{wu2025difix3d}, GSFixer~\cite{yin2025gsfixer}), our large-scale data scaling and homogeneous video diffusion framework ensure spatio-temporally consistent, high-fidelity restoration.}
  \label{fig:teaser}
\end{teaserfigure}


\maketitle

\section{Introduction}
\label{sec:intro}

Deep learning has advanced 2D representation and multimodal perception~\cite{xia2024vit,lou2025llava,shi2023open,li2026cadtrack,li2026ragtrack,kong2023dusa}, as well as generalizable 3D understanding and generation~\cite{xing2025doremi,kang2025sat2realcity,jia2026panoworld,li2026pano2world}. 3D Gaussian Splatting (3DGS) has attracted attention for real-time, high-fidelity rendering~\cite{kerbl20233dgs,sun2024f,xing2026adaptsplat,xu2025cruise,jia2026you}. However, the performance of 3DGS relies heavily on dense view coverage. Under sparse-view constraints, the optimization process often falls into a severely under-constrained state~\cite{zhu2024fsgs}. This limitation leads to geometric and photometric degradations in rendered views, such as floaters, background dilation, needle-like artifacts, and color outliers caused by the overfitting of Spherical Harmonics. These defects significantly hinder the practical industrial deployment of 3DGS in scenarios with limited capture.

To mitigate these issues, recent research utilizes large-scale generative models as diffusion priors to guide restoration~\cite{wu2024reconfusion,wu2025genfusion}. However, these paradigms face several core challenges. First, certain methods perform per-frame restoration, causing multi-view inconsistencies like flickering during novel view synthesis~\cite{wu2025difix3d}. Second, existing frameworks suffer from severe training data scarcity, making it difficult to generalize to diverse real-world artifact distributions. Furthermore, relying on external feature extractors~\cite{yin2025gsfixer} often introduces feature space mismatches and increases optimization difficulty. Critically, directly applying powerful diffusion priors without explicit spatial constraints causes models to misinterpret highly coupled 3DGS artifacts as standard 2D noise, triggering hallucinations and erroneous restorations (as illustrated in Figure~\ref{fig:visualizaiton_2d}).

In this paper, we present ArtifactWorld, a comprehensive framework for 3DGS artifact restoration through systematic data expansion and a homogeneous dual-model architecture. To address the data bottleneck, we first establish a fine-grained phenomenological taxonomy encompassing nine 3DGS degradation types. By leveraging an automated generative data flywheel, we expand our training set to 107,520 diverse paired video clips, significantly enhancing the model's robustness against complex artifacts.

Architecturally, we propose a homogeneous dual-model paradigm unifying the restoration process within a video diffusion backbone. Sharing this latent space, an isomorphic predictor first generates an explicit artifact heatmap. To resolve erroneous restorations, our Artifact-Aware Triplet Fusion (AATF) mechanism utilizes this heatmap as spatial intensity guidance to dynamically modulate the restoration strategy. In severely corrupted regions, it stimulates stronger generative capacity, synergizing with semantic cues from degraded reference frames to rebuild structures. Conversely, in low- or no-artifact areas, a reference mechanism directly extracts reliable features from clean boundary ground-truths introduced by Decoupled Boundary Anchoring (DBA) or reference frames. Coupled with a piecewise heatmap decay strategy, the model dynamically balances global context absorption and precise local repair. Finally, pristine frames are distilled back into 3D space via generative closed-loop reconstruction, eliminating 3DGS geometric defects. Experiments show ArtifactWorld excels in sparse-view synthesis and 3D reconstruction, yielding superior visual fidelity over existing methods (Figure~\ref{fig:teaser}). Our main contributions are:
\begin{itemize}[leftmargin=*,noitemsep,topsep=0pt,parsep=0pt]
    \item We introduce a phenomenological taxonomy of 3DGS sparse-view degradations and construct the ArtifactWorld Benchmark to facilitate standardized evaluation.
    \item We develop an automated generative data flywheel and construct a large-scale dataset of 107.5K paired video clips, establishing a foundation for learning degradation distributions.
    \item We design a homogeneous restoration network featuring Artifact-Aware Triplet Fusion (AATF) and Decoupled Boundary Anchoring (DBA), utilizing spatial artifact intensity guidance to achieve precise spatio-temporal repair within native self-attention layers.
    \item Extensive experiments validate that our framework achieves state-of-the-art performance in both 2D artifact restoration and 3D reconstruction tasks, demonstrating superior robustness across various sparsity protocols.
\end{itemize}

\section{Related Work}
\label{sec:related}

\subsection{Regularization-based Sparse-View 3DGS}
To mitigate sparse-view 3DGS degradations (e.g., floaters and background dilation), previous studies have primarily focused on per-scene regularizations, such as monocular depth priors~\cite{chung2024depth,zhu2024fsgs}, frequency smoothness~\cite{zhang2024fregs}, dropout strategies~\cite{park2025dropgaussian,xu2025dropoutgs} or explicit geometric and denoising constraints~\cite{zhang2024cor,xu2026denoise}. However, limited by their reliance on existing geometric cues, these analytical approaches struggle to hallucinate missing high-frequency details in severely under-observed regions, typically yielding overly smoothed results that remain sensitive to specific scene distributions.

\begin{figure*}[t]
    \centering
    \includegraphics[width=0.93\linewidth]{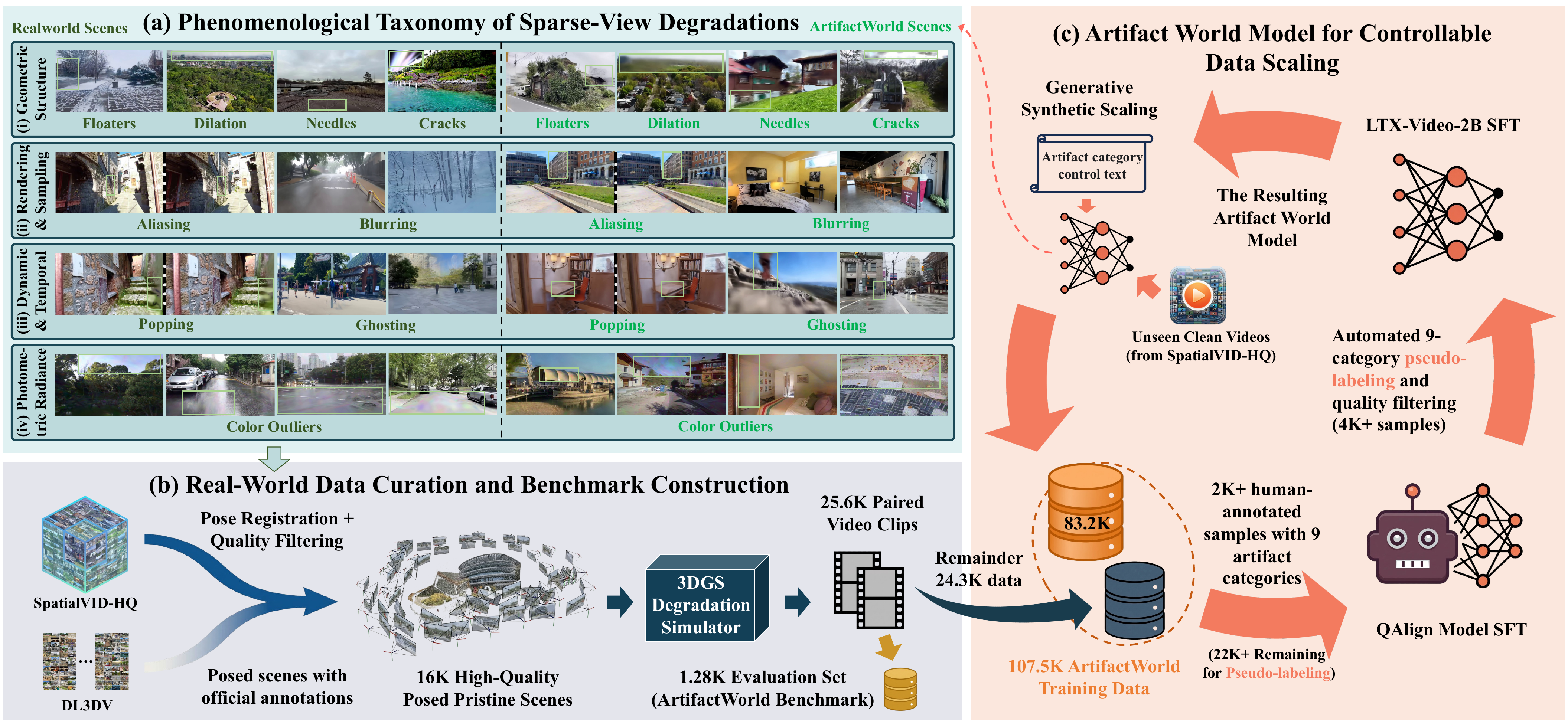}
    \vspace{-2mm}
    \caption{\textbf{Overview of ArtifactWorld Data Engine.} (a) Our phenomenological taxonomy categorizes 3DGS sparse-view degradations into 4 domains and 9 artifact types. (b) The data curation pipeline extracts 16K pristine scenes and physically simulates 25.6K paired videos, splitting them into a 1.28K benchmark and 24.3K training data. (c) The generative data flywheel leverages a VLM and an Artifact World Model to ultimately produce 107.5K training pairs.}
    \label{fig:visualizaiton}
    \vspace{-2mm}
\end{figure*}
\vspace{-1mm}
\subsection{Generative Priors for 3D Restoration}
Generative priors have emerged to overcome the limitations of analytical regularization. For NeRFs, ReconFusion~\cite{wu2024reconfusion} uses image diffusion to optimize novel views; for 3DGS, 3DGS-Enhancer~\cite{liu20243dgsenhancer}, Difix3D~\cite{wu2025difix3d}, GenFusion~\cite{wu2025genfusion}, and GSFixer~\cite{yin2025gsfixer} use diffusion priors to correct rendering artifacts. However, two challenges remain. First, image-based distillation lacks temporal awareness, causing cross-view flickering and popping. Second, although video-based approaches improve temporal coherence, they often attach external feature extractors (e.g., DINOv2~\cite{oquab2023dinov2} and VGGT~\cite{wang2025vggt}) to diffusion backbones via cross-attention, increasing computation and inducing feature-space mismatch. ArtifactWorld instead performs spatio-temporal restoration within the native self-attention of a homogeneous video diffusion backbone, avoiding external tokenizers and enabling large-scale data training.
\vspace{-1mm}
\subsection{Video Generation and Restoration}
Foundation video diffusion models (DiTs) excel in modeling spatio-temporal dynamics and in-context generation~\cite{yangcogvideox,geyertokenflow,mei2025field,yin2025slow,cai2025ditctrl,qi2025mask,blattmann2023stable}. Recently, these generative priors have advanced complex video restoration tasks~\cite{xie2025star,chen2025dove,wang2025seedvr2,zhuang2025flashvsr,xu2025videogigagan,zhou2024upscale}. However, directly applying generic video restoration paradigms to sparse-view 3DGS is challenging. General priors blindly treat highly coupled 3DGS artifacts as standard 2D noise, causing severe geometric over-hallucination and degrading accurately reconstructed regions. To resolve this bottleneck, ArtifactWorld employs an isomorphic predictor to generate explicit spatial heatmaps. This adaptive conditioning strictly restricts generative modifications to corrupted regions, preserving the multi-view consistency of healthy 3DGS geometry.

\section{ArtifactWorld Benchmark \& Data Scaling}
\label{sec:data}

\subsection{Phenomenological Taxonomy of Degradations}
\label{sec:taxonomy}

To systematically categorize the failure modes of 3DGS under sparse constraints, considering that the attributes in the 3DGS optimization process are highly coupled---making it extremely difficult to completely decouple the underlying causes of degradation---our taxonomy avoids exhaustive causal tracing. Instead, we adopt a phenomenological perspective from four fundamental representation domains, as shown in Figure~\ref{fig:visualizaiton}(a):

\noindent\textbf{(i) Geometric Structure:}
\begin{itemize}
[leftmargin=*,noitemsep,topsep=0pt,parsep=0pt]
    \item \textbf{Floaters:} High-opacity primitives wandering in free space, typically induced by overfitting multi-view consistency discrepancies due to a lack of strict 3D geometric constraints.
    \item \textbf{Dilation:} Amorphous, oversized Gaussian blobs. This stems from insufficient densification, forcing primitives to over-expand to cover large areas to minimize photometric loss, particularly in textureless regions or under severe depth ambiguity.
    \item \textbf{Needles:} Sharp, spike-like artifacts emerging when Gaussians become highly anisotropic along specific rays to overfit sparse views. They manifest prominently from unconstrained novel perspectives (e.g., close-up or orthogonal).
    \item \textbf{Cracks:} Manifest as topological voids appearing on object surfaces or structures. The potential underlying cause is primarily the lack of explicit topological constraints, causing overlapping Gaussian spheres to experience spatial positional drift.
\end{itemize}

\noindent\textbf{(ii) Rendering \& Sampling:}
\begin{itemize}
[leftmargin=*,noitemsep,topsep=0pt,parsep=0pt]
    \item \textbf{Aliasing:} High-frequency Moiré patterns. Critical sub-pixel aliasing occurs when rendering distant views of models optimized on close-up views; without low-pass pre-filtering, pixel grids severely undersample shrunken 2D Gaussian footprints.
    \item \textbf{Blurring:} Overly smoothed textures. When densification limits or multi-view inconsistencies prevent spawning high-frequency primitives, the optimizer over-expands Gaussian scales, functionally degenerating them into spatial low-pass filters.
\end{itemize}

\noindent\textbf{(iii) Dynamic \& Temporal:}
\begin{itemize}
[leftmargin=*,noitemsep,topsep=0pt,parsep=0pt]
    \item \textbf{Popping:} Rendered results exhibit discontinuous flickering or jumping during camera motion. The common cause is that intersecting Gaussian spheres in space undergo irregular sorting order exchanges as the camera moves.
    \item \textbf{Ghosting:} Semi-transparent, trailing, or duplicate artifacts. As dynamic elements violate static assumption, optimizer lowers opacity to minimize cross-view photometric errors, forcefully blending transient objects into the static background.
\end{itemize}

\noindent\textbf{(iv) Photometric Radiance:}
\begin{itemize}
[leftmargin=*,noitemsep,topsep=0pt,parsep=0pt]
    \item \textbf{Color Outliers:} Erratic color shifts or noise under novel views. These artifacts arise from unregularized high-degree SH overfitting to limited training rays and are amplified by imperfect multi-view exposure compensation or storage quantization.
\end{itemize}
\vspace{-3mm}
\subsection{Data Curation and Benchmark Construction}
\label{sec:data_curation}
\noindent\textbf{Source Data Mining \& Pose Registration.} 
To build a taxonomy-aligned dataset, our source pool integrates 10K raw scenes with pose annotations from DL3DV~\cite{ling2024dl3dv} and 118K rigid sequences mined from SpatialVID-HQ~\cite{wang2025spatialvid} using a dynamic ratio threshold ($<0.01$). To ensure high-quality reconstruction, we utilize an enhanced DROID-SLAM~\cite{teed2021droid} with Depth-Anything V2~\cite{yang2024depth} priors, featuring UniDepth~\cite{piccinelli2024unidepth} initialization, decoupled two-stage tracking, and geometric consistency filtering to prune high reprojection errors.

\noindent\textbf{Physics-Grounded Trajectory Quality Filtering.} 
Standard tracking metrics often fail to capture high-frequency jitters severely degrading novel view synthesis. Therefore, we introduce a physics-grounded smoothness filter enforcing translational (jerk), rotational (angular acceleration), and directional consistency constraints. Using Median Absolute Deviation (MAD), we robustly prune violating frames to distill \textbf{16,043} high-quality, registered static scenes.
\begin{figure*}[t]
    \centering
    \includegraphics[width=0.85\linewidth]{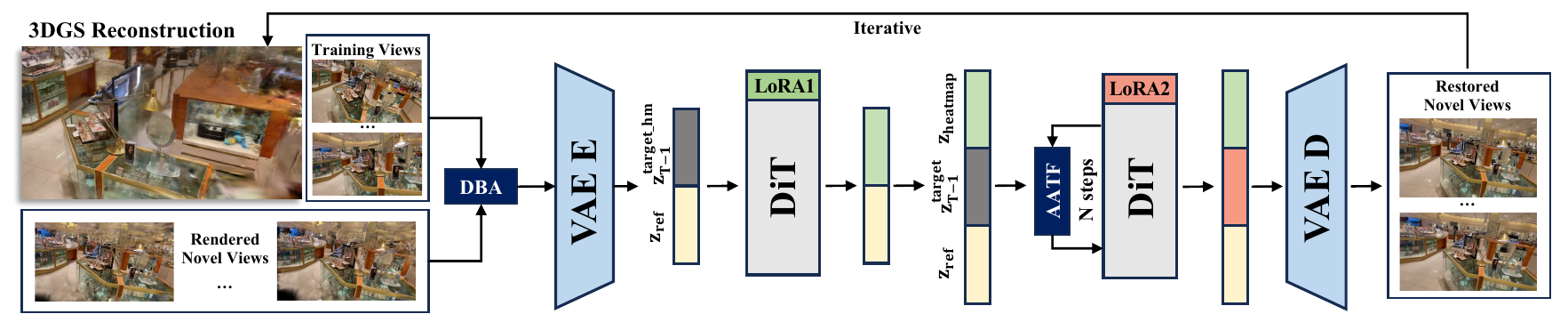}
    \vspace{-3mm}
    \caption{\textbf{ArtifactWorld Framework.} Under a Homogeneous Dual-Model Paradigm within a diffusion transformer: (1) \textbf{DBA} extracts clean boundary references. (2) A predictor (optimized and inferred via \textbf{LoRA1}) generates artifact heatmaps. (3) A restoration model (optimized and inferred via \textbf{LoRA2}) leverages these heatmaps to synthesize high-fidelity frames via \textbf{AATF}. (4) Closed-loop optimization uses these frames to permanently eliminate 3DGS artifacts.}
    \label{fig:visualizaiton_method}
    \vspace{-3mm}
\end{figure*}

\noindent\textbf{Physics-Grounded Degradation Simulation.} 
Based on 16K pristine scenes, we physically synthesize \textbf{25,616} paired clips via 3DGS rasterization replicating nine artifact categories: (1) Sparsity \& Optimization Stage Intervention: extreme sparsity (e.g., $8\times$ frame skipping) induces \textit{Needles, Floaters}, and \textit{Popping} during late-stage overfitting, while early-stage underfitting or dynamic objects trigger \textit{Blurring} and \textit{Ghosting}. (2) Random Checkpoint Sampling (1.5K--12K iterations) captures the temporal evolution of degradations. (3) Explicit Parameter Perturbations: we induce \textit{Cracks} via logarithmic scale compression, exacerbate \textit{Dilation} via 20\% point dropout, and trigger \textit{Color Outliers} by injecting Gaussian noise into SH coefficients, with resolution downsampling synthesizing \textit{Aliasing}.

\noindent\textbf{The ArtifactWorld Benchmark.} 
From 25,616 pairs, we partition \textbf{24,332} clips for training and \textbf{1,284} for the golden \textit{ArtifactWorld Benchmark}. This manually audited test set ensures a balanced distribution of all nine artifact categories, providing the most comprehensive, rigorous benchmark for 3DGS restoration to date.

\subsection{Artifact World Model for Data Scaling}
\label{sec:world_model}
Relying solely on the 3DGS rasterization pipeline to physically synthesize hundred-thousand-scale degraded videos is not only computationally prohibitive but also inflexible for specifying complex artifact combinations. To break this scaling bottleneck, we introduce a text-conditioned diffusion prior, designing a generative data flywheel (Figure~\ref{fig:visualizaiton}c). First, by employing a VQA strategy with balanced positive and negative samples, we fine-tune the Q-Align~\cite{wu2024q} model on \textbf{2,285} precisely annotated videos. Achieving a QA-level accuracy of 94.56\% and an F1 score of 0.89, this VLM serves as a strict gatekeeper to distill \textbf{4,385} pairs of high-confidence pseudo-labeled samples from the physically simulated data. Subsequently, utilizing this 4K dataset, we fine-tune LTX-Video-2B to construct a controllable Artifact World Model. We introduce a combinatorial prompting strategy (randomly sampling 1 to 9 artifact categories and shuffling them) to guide this model in learning the mapping from clean geometry to complex degradation patterns. By applying this generative prior to unseen videos from SpatialVID-HQ, we synthesize \textbf{83,188} multi-degradation clips. Merging these with the simulated data ultimately constructs a large-scale training set of \textbf{107,520} samples, thoroughly unlocking the data scaling capabilities for 3DGS restoration. Further implementation details are provided in the Supplementary Material.

\section{Methodology}
\label{sec:method}
\subsection{Framework Overview}
Our ArtifactWorld framework operates on a Flow Matching video diffusion model (LTX-Video~\cite{hacohen2024ltx}). Given a video $V$, a pre-trained VAE encodes it into a compact latent $z = \mathcal{E}(V)$. The model learns to predict a velocity field $v_\theta$ that constructs an ODE path from standard Gaussian noise $z_1 \sim \mathcal{N}(0, I)$ to the data distribution $z_0$. 

To rectify sparse-view 3DGS artifacts, we propose a Homogeneous Dual-Model Paradigm, unifying restoration within native self-attention layers (Figure~\ref{fig:visualizaiton_method}). The coupled pipeline comprises: Isomorphic Artifact Prediction for spatial intensity guidance and Heatmap-Guided Video Restoration for intensity-adaptive feature routing (Sec.~\ref{sec:4_3}), alongside Closed-Loop Generative Reconstruction (Sec.~\ref{sec:4_4}) to permanently eliminate 3DGS geometric defects.

\subsection{Decoupled Boundary Anchoring (DBA)}
For video-to-video generation constrained by reference frames, native models often couple clean and noisy tokens within the same sequence, increasing optimization difficulty. To resolve this for both our predictor and restoration models, we propose Decoupled Boundary Anchoring (DBA). Instead of intra-sequence mixing, we isolate all conditional information into a highly structured reference latent sequence $\mathbf{z}_{ref}$, anchored exclusively by clean GT sparse views at its temporal boundaries. The target latent $\mathbf{z}_{t}^{target}$ is strictly maintained as a homogeneous generative sequence. Specifically, the reference sequence is formulated as:
\begin{equation}
    \mathbf{z}_{ref}[i] = 
    \begin{cases}
        \mathcal{E}(\mathbf{V}_{GT}[0]), & i = 0 \\
        \mathcal{E}(\mathbf{V}_{GT}[-1]), & i \in [T-k, T-1] \\
        \mathcal{E}(\mathbf{V}_{artifact}[i]), & \text{otherwise}
    \end{cases}
\end{equation}
where $k$ denotes the trailing anchor window size. Tailored to the asymmetric temporal compression of the LTX-Video latent space, we set $k=8$ to form a complete compressed chunk, ensuring robust terminal anchoring without introducing latent alignment bias. Through this decoupled formulation, the global self-attention mechanism can bidirectionally query these pristine spatial anchors.

\subsection{Homogeneous Dual-Model Paradigm}
\label{sec:4_3}
Based on the shared DBA formulation, we unify artifact localization and restoration into a Homogeneous Dual-Model Paradigm using the same LTX-Video-13B backbone and Flow Matching objective.

\noindent\textbf{Phase 1: Homogeneous Heatmap Predictor.}
Precise restoration demands pixel-level awareness of degraded regions. Instead of relying on external feature extractors, we reframe artifact localization as a homogeneous video generation task. The predictor takes a concatenated \textbf{doublet} as input: $\mathbf{z}_{input}^{pred} = \text{Concat}([\mathbf{z}_{ref}, \mathbf{z}_{t}^{target\_hm}])$, where $z_{t}^{target\_hm}$ denotes the heatmap-specific noisy latent. Conditioned on the DBA-anchored reference, the model is fine-tuned via the standard flow matching objective (instantiated as a lightweight adapter, denoted as LoRA1) to predict an explicit artifact heatmap. To train this predictor, we offline-generate pseudo-GT heatmaps by computing the multi-layer perceptual feature discrepancies (LPIPS) between artifact-corrupted and clean videos. Unlike low-level pixel-wise metrics (e.g., MSE) that often overlook structural defects, LPIPS leverages deep feature spaces to accurately capture human-perceptible 3DGS artifacts.

\noindent\textbf{Phase 2: Artifact-Aware Triplet Fusion.}
With the explicitly predicted heatmap serving as a native spatial prior, the restoration model (parameterized by a separate adapter, LoRA2, sharing the same frozen backbone) seamlessly extends the formulation into an Artifact-Aware Triplet Fusion (AATF). The input evolves into a homogeneous \textbf{triplet}: $\mathbf{z}_{input}^{rest} = \text{Concat}([\mathbf{z}_{ref}, \mathbf{z}_{t}^{target}, \mathbf{z}_{heatmap}])$. 
Aligned with the generative nature of Flow Matching, we design a Piecewise Heatmap Decay strategy. This approach dynamically modulates the spatial mask to control the network's utilization of different information dimensions within $\mathbf{z}_{ref}$. By defining two empirical thresholds $\tau_1$ and $\tau_2$ ($0 < \tau_1 < \tau_2 < 1$), the static mask is transformed into a time-dependent $\tilde{\mathbf{z}}_{heatmap}(t)$:
\begin{equation}
\tilde{\mathbf{z}}_{heatmap}(t) = 
\begin{cases} 
\alpha(t) \mathbf{z}_{full} + (1-\alpha(t)) \mathbf{z}_{heatmap}, & t \in [\tau_2, 1] \\ 
\mathbf{z}_{heatmap}, & t \in [\tau_1, \tau_2) \\ 
\beta(t) \mathbf{z}_{heatmap} + (1-\beta(t)) \mathbf{z}_{null}, & t \in [0, \tau_1) 
\end{cases}
\end{equation}
where $\mathbf{z}_{full}$ denotes an all-white mask (set to 1 except for boundary frames), representing the opening of global reconstruction permission; $\mathbf{z}_{null}$ denotes an all-black mask (set to 0), representing the closure of explicit reconstruction permission; $\alpha(t)$ and $\beta(t)$ are monotonic interpolation functions. This scheduling scheme coordinates the guiding role of $\mathbf{z}_{ref}$ through three steps:

Firstly, during the initial integration stage ($t \in [\tau_2, 1]$), the global permission provided by $\mathbf{z}_{full}$ allows the model to absorb color and style baselines from $\mathbf{z}_{ref}$ while avoiding being rigidly anchored to erroneous local geometries caused by artifacts in the intermediate frames, thereby establishing a coherent global topology. Then, during mid-stage denoising ($t \in [\tau_1, \tau_2)$), the mask reverts to the $\mathbf{z}_{heatmap}$ constraint. This strictly anchors the generation process to the artifact-free regions of $\mathbf{z}_{ref}$, utilizing these high-confidence local structures as references while concentrating computational effort on repairing damaged areas. Finally, in the late low-noise stage ($t \in [0, \tau_1)$), the mask decays toward $\mathbf{z}_{null}$ to close explicit modification commands. The model ceases structural intervention and instead extracts authentic high-frequency textures from $\mathbf{z}_{ref}$ as detail baselines. Leveraging the DiT prior, the model achieves adaptive feature alignment and seamless blending between restored and anchored regions, significantly improving perceptual quality.

The final triplet input, integrated with dynamic scheduling, is updated as $\mathbf{z}_{input}(t) = \text{Concat}([\mathbf{z}_{ref}, \mathbf{z}_{t}^{target}, \tilde{\mathbf{z}}_{heatmap}(t)])$.

\noindent\textbf{Masked Flow Matching Objective.}
To physically realize the aforementioned decoupled paradigm—which isolates the reference, heatmap, and target latent spaces—we apply a sequence-specific timestep assignment strategy during training. Given that the DiT architecture relies on timesteps to perceive the noise level of each token, the pristine reference tokens ($\mathbf{z}_{ref}$) and dynamic heatmap tokens ($\tilde{\mathbf{z}}_{heatmap}$) are statically assigned a timestep of $0$, firmly anchoring them as clean contexts. Conversely, the target generation sequence receives a uniformly sampled diffusion timestep $t \in (0, 1]$. Consequently, the dense timestep tensor for the full sequence is formulated as $\mathbf{t}_{seq} = [\mathbf{0}, t \cdot \mathbf{1}, \mathbf{0}]$.

Under this design, the optimization objective is strictly confined to the target generation space. Let $\mathbf{v}_\theta^{target}$ denote the sequence slice of the network's predicted velocity field corresponding exclusively to the target tokens, and $\Delta \mathbf{z} = \mathbf{z}_1^{target} - \mathbf{z}_0^{target}$ represent the target velocity. The final restoration loss, denoted as $\mathcal{L}_r$ for brevity, is computed via a Masked Flow Matching objective:
\begin{equation}
\mathcal{L}_r = \mathbb{E}_{t, \mathbf{z}_0, \mathbf{z}_1} \left[ \left\| \mathbf{v}_\theta^{target}(\mathbf{z}_{input}(t), \mathbf{t}_{seq}) - \Delta \mathbf{z} \right\|_2^2 \right]
\end{equation}
where $\mathbf{z}_1^{target} \sim \mathcal{N}(0, \mathbf{I})$ represents the sampled Gaussian noise, and $\mathbf{z}_0^{target}$ is the clean ground-truth (GT) target latent. By explicitly slicing the loss computation, we ensure that gradients are backpropagated exclusively through the generative denoising task. This effectively isolates the generation process from the conditional spatial anchors at the optimization level, preventing potential entanglement during training.

\subsection{Closed-Loop Generative Reconstruction}
\label{sec:4_4}
Following the restoration of the 2D frames, we employ an iterative generative reconstruction process to optimize the 3DGS representation. To ensure both rendering quality and comprehensive angular coverage, we adopt the Reference-guided Trajectory sampling strategy introduced by GSFixer~\cite{yin2025gsfixer}. Specifically, the camera trajectory is formulated such that its start and end frames strictly correspond to two given sparse training views (reference views). The intermediate trajectory is constructed by: (i) interpolating from the starting reference view to its nearest viewpoint on a spherical path, (ii) sampling intermediate novel views along this spherical path, and (iii) interpolating to the ending reference view. 

We render artifact-prone novel views along these trajectories and feed them into our proposed restoration model to obtain clean, artifact-free frames. These restored frames are then integrated into the training set to supervise the 3DGS optimization iteratively. During optimization, we freeze the video restoration model and update the 3DGS parameters using a combined loss function $\mathcal{L}$:
\begin{equation}
\mathcal{L} = \mathcal{L}_{recon} + \lambda \cdot \mathcal{L}_{gen}
\end{equation}
where $\mathcal{L}_{recon}$ is the reconstruction loss supervised by the original sparse views, and $\mathcal{L}_{gen}$ is the generative loss supervised by restored novel views. Both utilize standard photometric loss:
\begin{equation}
\mathcal{L}_{*} = \lambda_{l1} \cdot \mathcal{L}_{1} + \lambda_{SSIM} \cdot \mathcal{L}_{SSIM}
\end{equation}
where $\lambda_{l1}$ and $\lambda_{SSIM}$ are the loss weights. This straightforward distillation physically and permanently removes artifacts in the 3D space, completing our closed-loop framework.

\begin{figure*}[t]
    \centering
    \includegraphics[width=0.75\linewidth]{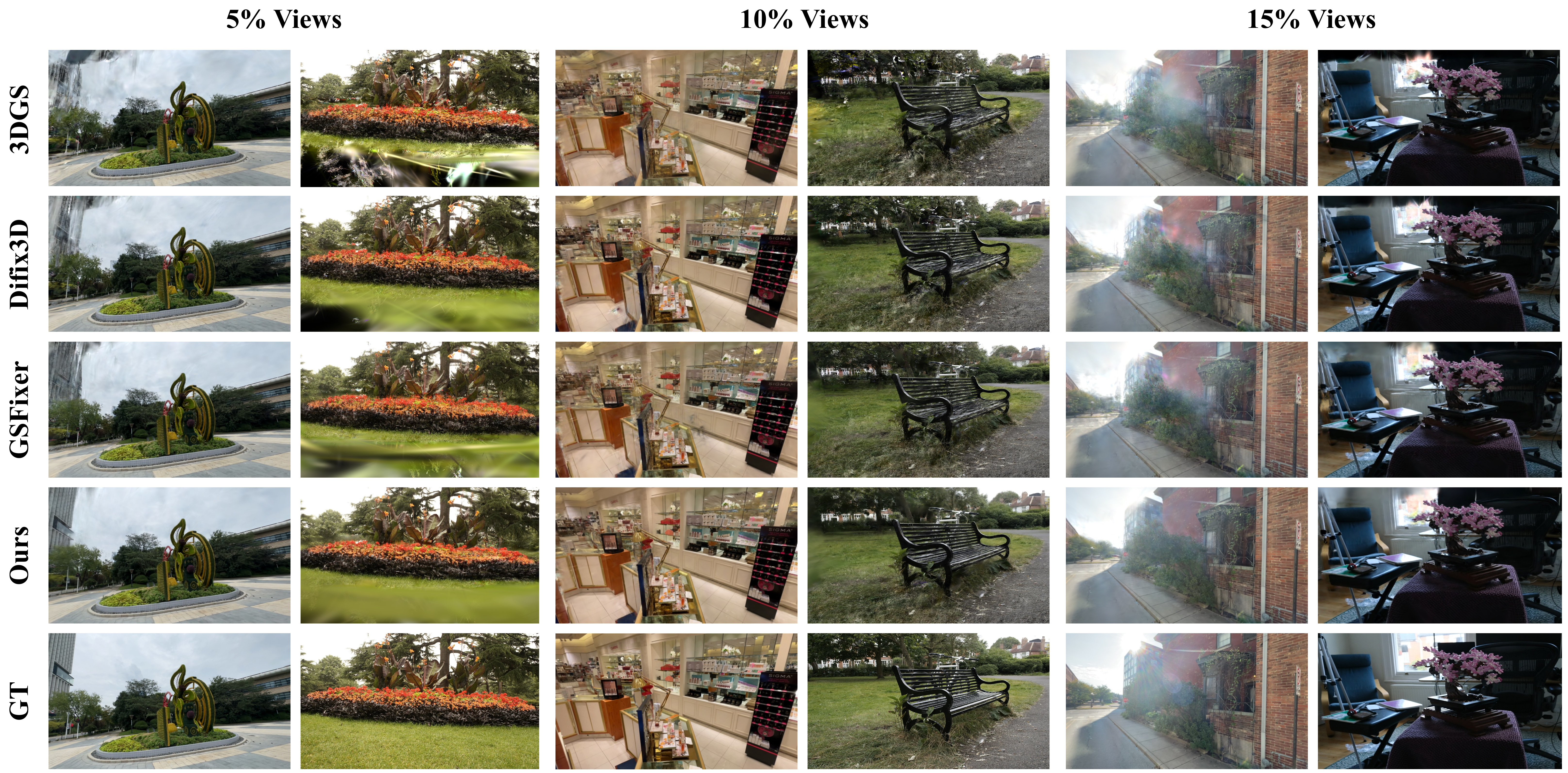}
    \vspace{-2mm}
    \caption{\textbf{Qualitative comparison of sparse-view 3D reconstruction.} Across varying sparsity ratios (5\%, 10\%, 15\%), ArtifactWorld consistently outperforms existing baselines by generating high-fidelity details closest to the ground truth.}
    \label{fig:visualizaiton_3d}
    \vspace{-1.5mm}
\end{figure*}
\begin{table*}[t]
  \centering
  \small
  \caption{Quantitative comparison on DL3DV and Mip-NeRF 360 datasets under different training view ratios. Best and second-best results are highlighted in \textbf{bold} and \underline{underline}, respectively. $^*$ denotes using the same camera trajectory as GSFixer.}
  \vspace{-2mm}
  \label{tab:comparison}
  \renewcommand{\arraystretch}{0.9}
  {%
  \begin{tabular}{ll ccc ccc ccc}
    \toprule
    \multirow{2}{*}{\textbf{Dataset}} & \multirow{2}{*}{\textbf{Method}}
      & \multicolumn{3}{c}{\textbf{PSNR$\uparrow$}}
      & \multicolumn{3}{c}{\textbf{SSIM$\uparrow$}}
      & \multicolumn{3}{c}{\textbf{LPIPS$\downarrow$}} \\
    \cmidrule(lr){3-5} \cmidrule(lr){6-8} \cmidrule(lr){9-11}
    & & \textbf{5\%} & \textbf{10\%} & \textbf{15\%} & \textbf{5\%} & \textbf{10\%} & \textbf{15\%} & \textbf{5\%} & \textbf{10\%} & \textbf{15\%} \\
    \midrule
    \multirow{4}{*}{DL3DV}
      & 3DGS~\cite{kerbl20233dgs}     & 22.75 & 26.17 & 27.76 & 0.7737 & 0.8583 & 0.8880 & 0.2247 & 0.1597 & 0.1372 \\
      & Difix3D$^*$~\cite{wu2025difix3d} & \underline{23.53} & \underline{26.57} & \underline{27.96} & \underline{0.7885} & \underline{0.8636} & \underline{0.8913} & \underline{0.2080} & \underline{0.1520} & \textbf{0.1315} \\
      & GSFixer~\cite{yin2025gsfixer}  & 23.44 & 26.45 & 27.93 & 0.7853 & 0.8615 & 0.8898 & 0.2161 & 0.1552 & \underline{0.1334} \\
      & ArtifactWorld~(Ours)     & \textbf{24.56} & \textbf{27.14} & \textbf{28.27} & \textbf{0.8111} & \textbf{0.8722} & \textbf{0.8942} & \textbf{0.1970} & \textbf{0.1498} & 0.1341 \\
    \midrule
    \multirow{4}{*}{Mip-NeRF-360}
      & 3DGS~\cite{kerbl20233dgs}     & 17.08 & 20.53 & 22.35 & 0.4533 & 0.5860 & 0.6440 & 0.4246 & 0.3192 & 0.2811 \\
      & Difix3D$^*$~\cite{wu2025difix3d} & 18.14 & 21.33 & \underline{22.92} & 0.4873 & 0.6036 & \underline{0.6545} & \textbf{0.4068} & \textbf{0.3078} & \textbf{0.2704} \\
      & GSFixer~\cite{yin2025gsfixer}  & \underline{18.73} & \underline{21.50} & 22.86 & \underline{0.4999} & \underline{0.6044} & 0.6521 & 0.4212 & 0.3158 & \underline{0.2761} \\
      & ArtifactWorld~(Ours)     & \textbf{19.97} & \textbf{22.56} & \textbf{23.79} & \textbf{0.5276} & \textbf{0.6274} & \textbf{0.6693} & \underline{0.4075} & \underline{0.3100} & 0.2773 \\
    \bottomrule
  \end{tabular}%
  }
  \vspace{-2mm}
\end{table*}
\section{Experiments}
\label{sec:experiments}

\subsection{Experimental Setup}
\label{sec:exp_setup}

\noindent\textbf{Datasets and Sparsity Protocol.} 
We evaluate on two widely adopted benchmarks: the in-domain DL3DV dataset~\cite{ling2024dl3dv} (sharing the same domain as the training set but with unseen scenes) and the out-of-domain generalization dataset Mip-NeRF 360~\cite{barron2022mipnerf360}, maintaining consistent scene configurations with GSFixer~\cite{yin2025gsfixer}. We employ a dual-sparsity protocol: first, a ratio-based protocol (uniformly sampling 5\%, 10\%, and 15\% of total frames) to ensure standardized and unbiased difficulty across scenes of diverse scales; second, an extreme sparsity evaluation using exactly 3 views on Mip-NeRF 360, following the baseline~\cite{yin2025gsfixer}, to validate robustness under highly restricted observations. Additionally, the 2D restoration capability is verified on the ArtifactWorld Benchmark, comprising 1,284 manually audited, taxonomy-aligned test clips.

\noindent\textbf{Evaluation Metrics.} 
For sparse-view 3D reconstruction, we report standard novel view synthesis metrics: PSNR, SSIM~\cite{wang2004image}, and LPIPS~\cite{zhang2018perceptual}. For 2D restoration, alongside the aforementioned metrics, we additionally introduce CLIP-I~\cite{radford2021learning} to evaluate semantic fidelity, and employ FVD~\cite{unterthiner2019fvd} to quantify spatio-temporal distributional discrepancies and temporal flickering.

\noindent\textbf{Implementation Details.} Our ArtifactWorld framework utilizes a frozen LTX-Video backbone with two task-specific adapters (LoRA1 for prediction and LoRA2 for restoration). Both adapters are fine-tuned via AdamW (lr=$3 \times 10^{-4}$, batch size 8, 14K iterations) on $8 \times$ A100 GPUs. The closed-loop 3DGS optimization strictly employs photometric losses ($\mathcal{L}_1$ and SSIM). Due to space constraints, exhaustive hyperparameters for the data pipeline and all models are relegated to the Supplementary Material.
\begin{figure*}[t]
    \centering
    \includegraphics[width=0.85\linewidth]{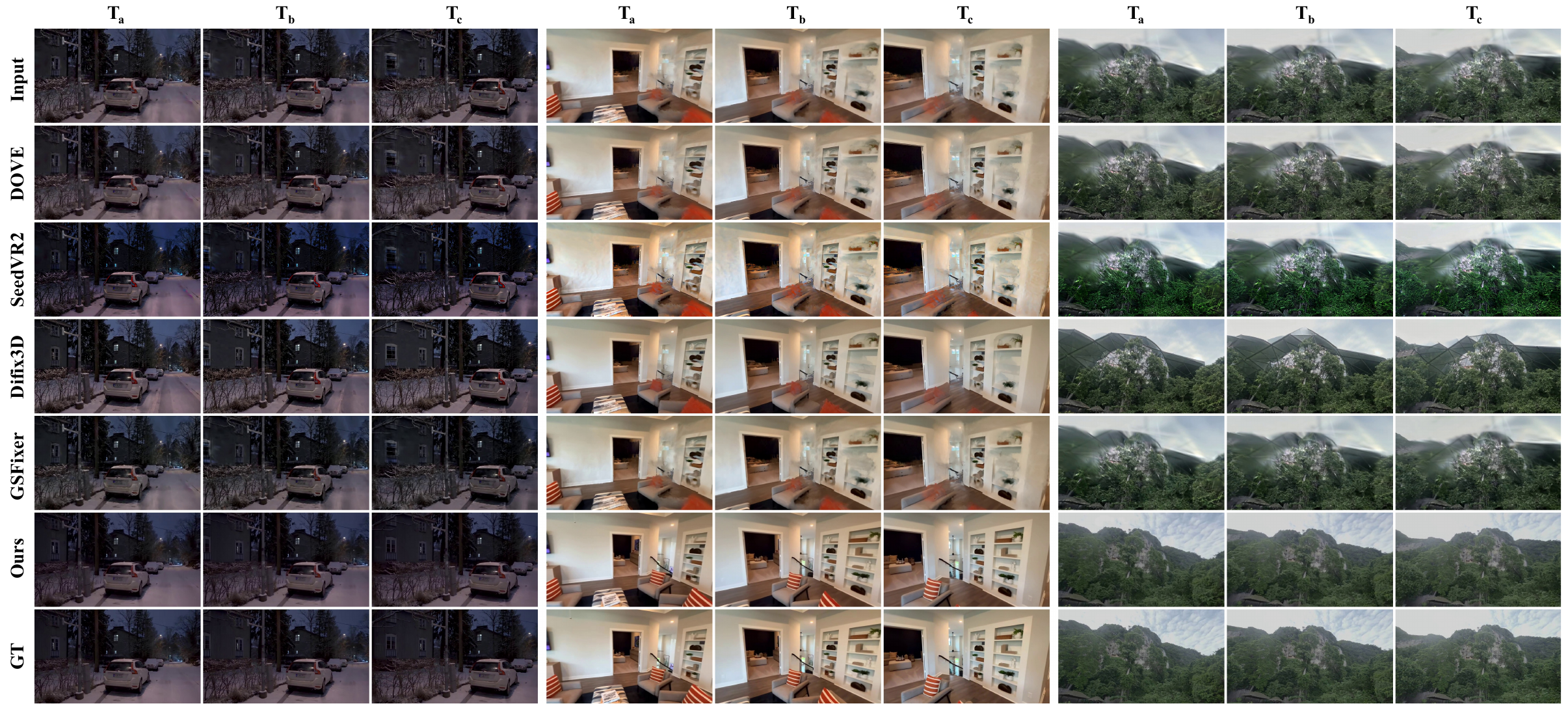}
    \vspace{-2mm}
    \caption{\textbf{Qualitative comparison of 2D artifact restoration.} Across sampled non-consecutive frames ($T_a < T_b < T_c$), ArtifactWorld visibly outperforms existing restoration baselines.}
    \label{fig:visualizaiton_2d}
    \vspace{-1.5mm}
\end{figure*}
\subsection{Sparse-View 3D Reconstruction}
\label{sec:exp_3d}
We compare our framework against recent generative baselines and vanilla 3DGS across various view ratios and scenes.

\noindent\textbf{Quantitative Analysis.}
Table~\ref{tab:comparison} shows that ArtifactWorld achieves state-of-the-art PSNR and SSIM across 5\%, 10\%, and 15\% view ratios on both in-domain DL3DV and out-of-domain Mip-NeRF 360, while remaining competitive in LPIPS. Notably, it delivers these gains on unseen Mip-NeRF 360 scenes without fine-tuning, demonstrating strong zero-shot generalization. Together, these results show that our data scaling and homogeneous architecture enable robust restoration across diverse indoor and outdoor scenes.

\noindent\textbf{Qualitative Analysis.}
Figure~\ref{fig:visualizaiton_3d} compares novel views on DL3DV (columns 1, 3, and 5) and Mip-NeRF 360 (columns 2, 4, and 6). Vanilla 3DGS produces severe geometric artifacts, including floaters and cracks, while Difix3D and GSFixer only partially correct heavily degraded regions. ArtifactWorld consistently yields the highest visual fidelity across both indoor and outdoor scenes.

\noindent\textbf{Resilience to Extreme Sparsity.}
To further explore the limits of generative reconstruction, we conduct a stress test under extreme sparsity using only 3 input views on the Mip-NeRF 360 dataset. As indicated by the metric comparisons in Table \ref{tab:sota_comparison_3d}, our method achieves the most robust novel view synthesis quality.

\noindent\textbf{Cross-Backbone Generalization.} We apply ArtifactWorld to Scaffold-GS~\cite{lu2024scaffold} on Mip-NeRF 360 without synthetic test-time degradation. ArtifactWorld consistently improves PSNR/SSIM at 5\% (18.01/0.4937 to 19.99/0.5496), 10\% (21.54/0.6196 to 22.86/0.6486), and 15\% (23.17/0.6678 to 24.09/0.6866) views. These gains confirm that ArtifactWorld generalizes beyond the original 3DGS backbone.

\subsection{2D Artifact Restoration Performance}
\label{sec:exp_2d}
\begin{table}[t]
  \centering
  \small
  \caption{Performance under extreme 3-view sparsity.}
  \vspace{-1.5mm}
  \label{tab:sota_comparison_3d}
  \renewcommand{\arraystretch}{0.9}
  \setlength{\tabcolsep}{10pt}
  \begin{tabular}{l ccc}
    \toprule
    \textbf{Method}       & \textbf{PSNR$\uparrow$} & \textbf{SSIM$\uparrow$} & \textbf{LPIPS$\downarrow$} \\
    \midrule
    SimpleNeRF\cite{somraj2023simplenerf}    & 13.27 & 0.283 & 0.741 \\
    ZeroNVS\cite{sargent2024zeronvs}       & 14.44 & 0.316 & 0.680 \\
    ReconFusion\cite{wu2024reconfusion}   & 15.50 & 0.358 & 0.585 \\
    3DGS\cite{kerbl20233dgs}          & 13.06 & 0.251 & 0.576 \\
    FSGS \cite{zhu2024fsgs}         & 14.17 & 0.318 & 0.578 \\
    Difix3D\cite{wu2025difix3d}      & 13.92 & 0.298 & 0.578 \\
    GenFusion\cite{wu2025genfusion}     & 15.03 & 0.357 & 0.578 \\
    GSFixer\cite{yin2025gsfixer}       & 15.61 & 0.370 & 0.559 \\
    ArtifactWorld~(Ours)          & \textbf{16.13} & \textbf{0.397} & \textbf{0.557} \\
    \bottomrule
  \end{tabular}
  \vspace{-1.5mm}
\end{table}

Beyond the final 3D representation, we isolate and evaluate the core capability of our framework: correcting complex 3DGS artifacts in the 2D video domain. On the benchmark comprising 1,284 clips, we comprehensively compare ArtifactWorld against two categories of state-of-the-art baselines: general video generation and restoration models (DOVE, SeedVR2) and concurrent generative 3D reconstruction pipelines (Difix3D, GSFixer).

\noindent\textbf{Qualitative Analysis.} Visually, as illustrated in Figure~\ref{fig:visualizaiton_2d}, general video restoration priors typically struggle to address the highly coupled artifacts specific to 3DGS, while previous 3D generative pipelines falter in temporal coherency or lack sufficient restoration strength. In contrast, ArtifactWorld precisely localizes and eliminates artifacts, recovering underlying high-frequency details while rigorously preserving the original scene structure.

\noindent\textbf{Quantitative Analysis.} As shown in Table~\ref{tab:comparison_2d}, ArtifactWorld demonstrates superiority across all metrics. Notably, its CLIP-I and FVD scores validate the method's effectiveness in maintaining semantic fidelity and spatio-temporal consistency.

\subsection{Ablation Studies and Empirical Analysis}
\label{sec:exp_ablation}

For efficiency, we perform ablation analysis on 200 randomly selected samples from the evaluation set.
\begin{table}[t]
  \centering
  \caption{Quantitative comparison of 2D artifact restoration.}
  \vspace{-1.5mm}
  \label{tab:comparison_2d}
  \renewcommand{\arraystretch}{0.9}
  \setlength{\tabcolsep}{4pt}
  \renewcommand{\arraystretch}{1.0}
  \small
  \begin{tabular}{lccccc}
    \toprule
    \textbf{Methods} & \textbf{PSNR$\uparrow$} & \textbf{SSIM$\uparrow$} & \textbf{LPIPS$\downarrow$} & \textbf{CLIP-I$\uparrow$} & \textbf{FVD$\downarrow$} \\
    \midrule
    DOVE~\cite{chen2025dove}       & 20.12 & 0.6529 & 0.4062 & 0.8423 & 174.05 \\
    SeedVR2~\cite{wang2025seedvr2} & 17.58 & 0.5840 & 0.4271 & 0.8232 & 197.94 \\
    Difix3D~\cite{wu2025difix3d}   & 20.21 & 0.6442 & 0.3617 & 0.8925 & 66.76  \\
    GSFixer~\cite{yin2025gsfixer}    & 20.19 & 0.6769 & 0.3834 & 0.8681 & 114.53 \\
    ArtifactWorld~(Ours)     & \textbf{24.20} & \textbf{0.7547} & \textbf{0.2893} & \textbf{0.9663} & \textbf{33.92} \\
    \bottomrule
  \end{tabular}
  \vspace{-4mm}
\end{table}

\noindent\textbf{Effectiveness of Architecture and Data Scaling.} As shown in Table~\ref{tab:ablation}, the native LTX-Video baseline (trained on 24K data) suffers from timestep feature coupling due to the direct concatenation of reference frames, which increases optimization difficulty (PSNR 23.44 dB, SSIM 0.7330). Introducing DBA eliminates this coupling and provides bidirectional spatial anchors, suppressing flickering and structural shifts (PSNR 23.99 dB, SSIM 0.7492). Integrating the AATF mechanism enables the network to leverage the heatmap to precisely target local corrupted regions, further improving the metrics to a PSNR of 24.15 dB and an SSIM of 0.7530. Finally, expanding to a diverse 107.5K dataset encompassing 9 artifact distributions achieves the optimal performance (PSNR 24.46 dB, LPIPS 0.2881). This confirms that data diversity improves robustness and generalization to complex real-world 3DGS degradations.

\begin{table}[t]
  \centering
  \small
  \caption{Ablation study of our proposed components.}
  \vspace{-1.5mm}
  \label{tab:ablation}
  \renewcommand{\arraystretch}{0.9}
  \begin{tabular}{@{}cccccc@{}}
    \toprule
    \textbf{DBA} & \textbf{AATF} & \textbf{Data Scaling} & \textbf{PSNR $\uparrow$} & \textbf{SSIM $\uparrow$} & \textbf{LPIPS $\downarrow$} \\
    \midrule
    $\times$ & $\times$ & $\times$ & 23.44 & 0.7330 & 0.3118 \\
    $\checkmark$ & $\times$ & $\times$ & 23.99 & 0.7492 & 0.2948 \\
    $\checkmark$ & $\checkmark$ & $\times$ & 24.15 & 0.7530 & 0.2900 \\
    $\checkmark$ & $\checkmark$ & $\checkmark$ & \textbf{24.46} & \textbf{0.7572} & \textbf{0.2881} \\
    \bottomrule
  \end{tabular}
  \vspace{-2mm}
\end{table}
\begin{figure}[t]
    \centering
    \includegraphics[width=0.9\linewidth]{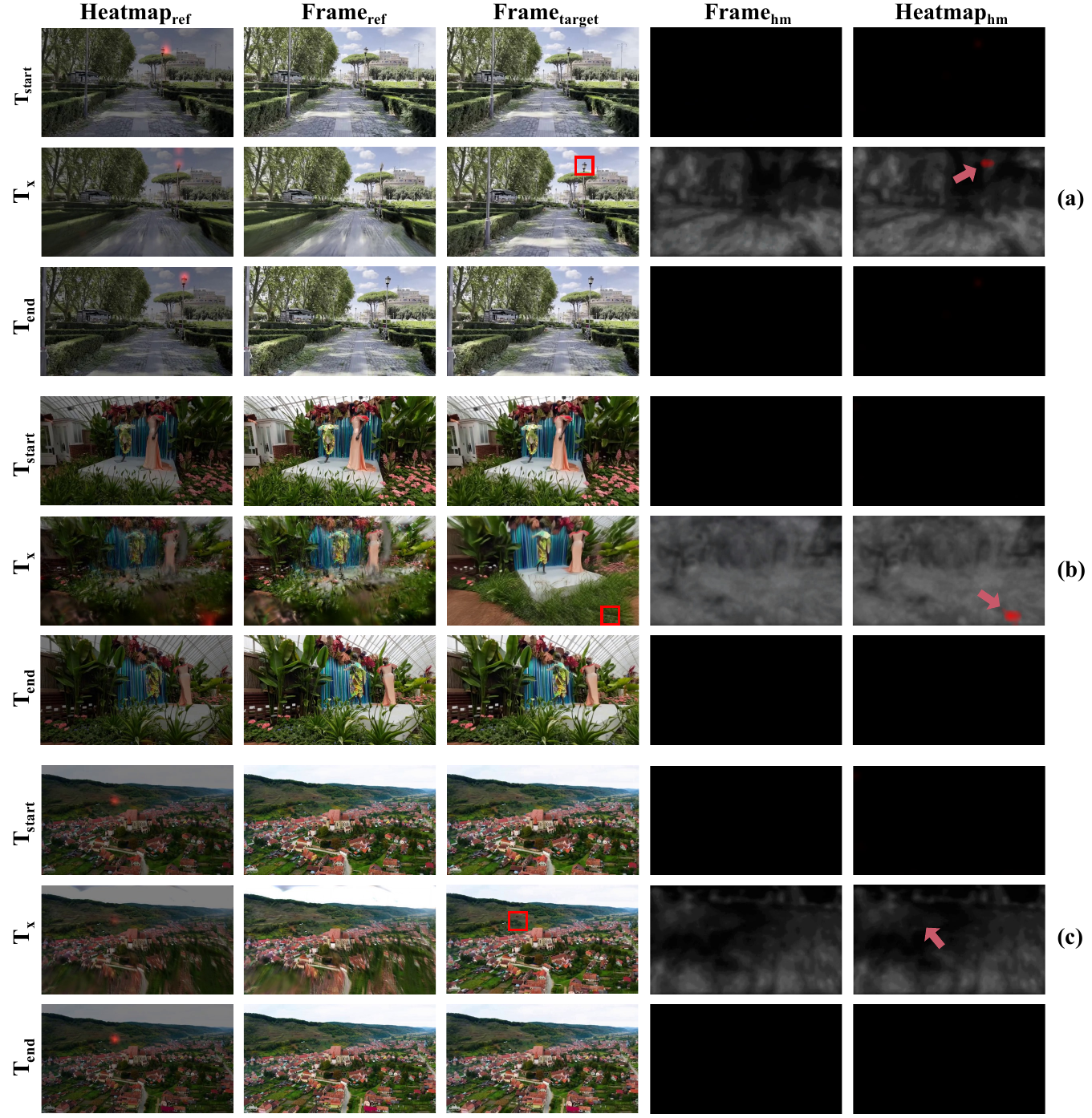}
    \vspace{-1.5mm}
    \caption{\textbf{Mechanistic Interpretability.} $T_{\text{start}}$, $T_x$, and $T_{\text{end}}$ denote the initial, intermediate, and terminal frames, respectively. $\text{Frame}_{\text{ref}}$, $\text{Frame}_{\text{target}}$ (visualized as the generated frame for clarity), and $\text{Frame}_{\text{hm}}$ represent the reference frame, target frame, and artifact heatmap, corresponding to latents $z_{\text{ref}}$, $z_t^{\text{target}}$, and $z_{\text{heatmap}}$. $\text{Heatmap}_{\text{ref}}$ and $\text{Heatmap}_{\text{hm}}$ visualize the spatial attention of the query token towards $z_{\text{ref}}$ and $z_{\text{heatmap}}$, respectively. (Zoom in for details.)}
    \label{fig:visualizaiton_hm}
    \vspace{-2mm}
\end{figure}

\noindent\textbf{Continuous Intensity Fidelity.} The Pearson Correlation Coefficient (PCC) between the predicted continuous heatmaps and LPIPS pseudo-GTs (Sec.~\ref{sec:4_3}) reaches 0.578 on the test set. In dense pixel-wise regression, this positive correlation confirms the network successfully captures the \textit{relative severity ranking} of 3DGS artifacts, providing a robust spatial prior for video restoration.

\noindent\textbf{Mechanistic Interpretability.} Figure 6 illustrates the attention distribution of query tokens in the target frame $T_x$ towards the reference ($z_{ref}$) and heatmap ($z_{heatmap}$) latents. \textbf{(1) Effectiveness of DBA:} Queries visible in the boundary frames (Cases a and c) exhibit strong activations at corresponding boundaries in $Heatmap_{ref}$, whereas occluded queries (Case b) show zero activation. This confirms that DBA precisely anchors clean temporal contexts, effectively avoiding feature hallucinations caused by spatial misalignments. \textbf{(2) Effectiveness of AATF:} The network's attention to the heatmap ($Heatmap_{hm}$) strictly correlates with artifact severity. Case (b), suffering from severe artifacts, assigns peak attention to the heatmap to stimulate strong generative capacity for rebuilding corrupted structures. Case (a), with only mild artifacts, extracts abundant information from the reference frames, thus exhibiting weak heatmap attention for subtle corrections. Conversely, the artifact-free Case (c) shows zero heatmap activation, relying entirely on reference frames to extract reliable features. This progression explicitly validates that AATF achieves intensity-adaptive feature routing based on artifact severity.

\begin{table}[t]
\centering
\caption{\textbf{Ablation on dynamic timestep scheduling.} Evaluated across an 8-step denoising process.}
\vspace{-1.5mm}
\label{tab:heatmap_decay}
\resizebox{\linewidth}{!}{
\begin{tabular}{l c c c c c c}
\toprule
\multirow{2}{*}{\textbf{Exp}} & \multicolumn{3}{c}{\textbf{Timestep Scheduling Strategy (Total: 8 Steps)}} & \multirow{2}{*}{\textbf{PSNR$\uparrow$}} & \multirow{2}{*}{\textbf{SSIM$\uparrow$}} & \multirow{2}{*}{\textbf{LPIPS$\downarrow$}} \\
\cmidrule(lr){2-4}
 & \textbf{Phase 1} ($t \ge \tau_2$) & \textbf{Phase 2} ($\tau_1 \le t < \tau_2$) & \textbf{Phase 3} ($t < \tau_1$) & & & \\
\midrule
1 & \multicolumn{3}{c}{Constant $z_{heatmap}$ (8 steps)} & 24.428 & 0.7565 & 0.2885 \\
2 & $z_{full}$ (1 step) & \multicolumn{2}{c}{$z_{heatmap}$ (7 steps)} & 24.433 & 0.7565 & 0.2883 \\
3 & \multicolumn{2}{c}{$z_{heatmap}$ (7 steps)} & $z_{null}$ (1 step) & 24.435 & 0.7566 & 0.2884 \\
4 & $z_{full}$ (1 step) & $z_{heatmap}$ (6 steps) & $z_{null}$ (1 step) & 24.439 & 0.7567 & 0.2882 \\
5 & $z_{null}$ (1 step) & $z_{heatmap}$ (6 steps) & $z_{full}$ (1 step) & 24.399 & 0.7566 & 0.2884 \\
6 & Linear to $z_{heatmap}$ (4 steps) & $z_{heatmap}$ (3 steps) & $z_{null}$ (1 step) & 24.402 & 0.7550 & 0.2899 \\
7 & $z_{full}$ (1 step) & $z_{heatmap}$ (3 steps) & Linear to $z_{null}$ (4 steps) & \textbf{24.459} & \textbf{0.7572} & \textbf{0.2881} \\
\bottomrule
\end{tabular}
}
\vspace{-2mm}
\end{table}

\noindent\textbf{Ablation on Timestep-Scheduled Heatmap Decay.} Table \ref{tab:heatmap_decay} ablates our dynamic scheduling across an 8-step denoising process. A constant $z_{heatmap}$ mask (Exp 1) proves sub-optimal. Introducing a 1-step initial $z_{full}$ (Exp 2) or a 1-step final $z_{null}$ (Exp 3) improves performance, yielding cumulative gains when combined (Exp 4). Reversing this order (Exp 5) or replacing the initial 1-step $z_{full}$ with a 4-step linear decay (Exp 6) degrades structural fidelity. Our optimal strategy (Exp 7) strictly maps to the proposed three-stage formulation to achieve the best restoration: a 1-step $z_{full}$ guidance in Phase 1, steady $z_{heatmap}$ anchoring for 3 steps in Phase 2, and a 4-step linear decay toward $z_{null}$ in Phase 3.

\subsection{Computational Footprint and Efficiency}
\label{sec:efficiency}

Table~\ref{tab:efficiency} summarizes per-scene runtime and peak memory on a single A100 under the DL3DV setting, together with restoration quality on Mip-NeRF 360 averaged over 5\%, 10\%, and 15\% views. Full ArtifactWorld achieves the best PSNR/SSIM while remaining more efficient than GSFixer. For latency-sensitive deployment, we introduce \textbf{ArtifactWorld-Turbo}, a training-free inference variant using fewer denoising steps and lower spatial resolution. It reduces closed-loop time from 113.3 to 48.8 min/scene (2.3$\times$ speedup) and memory from 46.3 to 41.6 GB, with only minor quality drops. ArtifactWorld-Turbo still outperforms both Difix3D and GSFixer in PSNR/SSIM despite its reduced computation, providing a practical efficiency-quality trade-off without retraining.

\begin{table}[t]
\centering
\scriptsize
\caption{Efficiency and restoration quality of generative reconstruction methods.}
\vspace{-1.5mm}
\label{tab:efficiency}
\setlength{\tabcolsep}{2.5pt}
\renewcommand{\arraystretch}{0.9}
\resizebox{\columnwidth}{!}{%
\begin{tabular}{@{}lcccc@{}}
\toprule
\textbf{Method}
& \textbf{Time (min)$\downarrow$}
& \textbf{Memory (GB)$\downarrow$}
& \textbf{PSNR$\uparrow$}
& \textbf{SSIM$\uparrow$} \\
\midrule
Difix3D~\cite{wu2025difix3d}
& \textbf{27.8}
& \textbf{13.6}
& 20.80
& 0.5818 \\
GSFixer~\cite{yin2025gsfixer}
& 146.4
& 50.8
& 21.03
& 0.5855 \\
ArtifactWorld
& 113.3
& 46.3
& \textbf{22.11}
& \textbf{0.6081} \\
ArtifactWorld-Turbo
& \underline{48.8}
& \underline{41.6}
& \underline{21.95}
& \underline{0.6002} \\
\bottomrule
\end{tabular}%
}
\vspace{-2mm}
\end{table}

\section{Conclusion}
\label{sec:conclusion}
We present ArtifactWorld, a scalable framework for resolving sparse-view 3D Gaussian Splatting degradations. To overcome data bottlenecks, we establish a phenomenological artifact taxonomy and synthesize a 107.5K video dataset via an automated generative flywheel. We propose a homogeneous dual-model paradigm that unifies restoration within native self-attention layers. Combining an isomorphic heatmap predictor with Decoupled Boundary Anchoring, we formulate restoration as an intensity-guided triplet interaction, enforcing explicit spatial constraints for precise structural repair and high spatio-temporal consistency. Extensive evaluations confirm state-of-the-art performance in both artifact correction and robust 3D reconstruction. Future work will explore cross-resolution generative priors to synthesize ultra-high-definition details for high-resolution scene reconstruction.

\bibliographystyle{ACM-Reference-Format}
\balance
\bibliography{sample-base}

\end{document}